\documentclass[letterpaper]{article} 
\usepackage{aaai25}  
\usepackage{times}  
\usepackage{helvet}  
\usepackage{courier}  
\usepackage[hyphens]{url}  
\usepackage{graphicx} 
\usepackage{natbib}  
\usepackage{caption} 
\usepackage{algorithm}
\usepackage{algorithmic}

\usepackage{newfloat}
\usepackage{listings}
\DeclareCaptionStyle{ruled}{labelfont=normalfont,labelsep=colon,strut=off} 
\floatstyle{ruled}
\newfloat{listing}{tb}{lst}{}
\floatname{listing}{Listing}
\usepackage{amsmath}
\usepackage{amsfonts}
\usepackage{subcaption}
\usepackage[frozencache,cachedir=.]{minted}
\usepackage{tcolorbox}
\usepackage{caption}
\usepackage{tabularx}
\usepackage{booktabs}

\newcommand{\subtwosection}[1]{\noindent\textbf{#1}}

\title{Enhance Modality Robustness in Text-Centric Multimodal Alignment with Adversarial Prompting}

\author {
    Yun-Da Tsai\textsuperscript{\rm 1},
    Ting-Yu Yen\textsuperscript{\rm 1},
    Keng-Te Liao,
    Shou-De Lin
}
\affiliations {
    National Taiwan University\\
    f08946007@csie.ntu.edu.tw, r11922042@ntu.edu.tw, d05922001@ntu.edu.tw, sdlin@csie.ntu.edu.tw
}

\usepackage{bibentry}

\begin{document}

\maketitle

\begin{abstract}
Converting different modalities into generalized text, which then serves as input prompts for large language models (LLMs), is a common approach for aligning multimodal models, particularly when pairwise data is limited. Text-centric alignment method leverages the unique properties of text as a modality space, transforming diverse inputs into a unified textual representation, thereby enabling downstream models to effectively interpret various modal inputs. This study evaluates the quality and robustness of multimodal representations in the face of noise imperfections, dynamic input order permutations, and missing modalities, revealing that current text-centric alignment methods can compromise downstream robustness. To address this issue, we propose a new text-centric adversarial training approach that significantly enhances robustness compared to traditional robust training methods and pre-trained multimodal foundation models. Our findings underscore the potential of this approach to improve the robustness and adaptability of multimodal representations, offering a promising solution for dynamic and real-world applications.
\end{abstract}

%

\section{Introduction}
\label{sec:intro}

Text-centric multimodal alignment methods have emerged as a powerful approach for integrating multimodal information by converting diverse data types into text. This technique leverages the unique properties of text as a universal modality space, enabling large language models (LLMs) to process and understand visual, auditory, and other forms of data, and have shown competetive performance compared to other traditional embedding-based alignment methods~\cite{tsai2024text}. By transforming non-textual information into textual descriptions, these methods facilitate the alignment and integration of various modalities, enhancing the capability of LLMs to comprehend and generate contextually rich responses.
For example, LLaVA~\cite{liu2023llava} uses expert models to generate captions, object detection locations, and textual descriptions from images. These are then used as input to GPT-4 to create vision-text instruction-following data as a substitute of actual collecting vision-text instruction-following data, which is inherently difficult and resource-intensive to obtain.

\begin{figure}[t]
    \centering
    \includegraphics[width=1.0\linewidth]{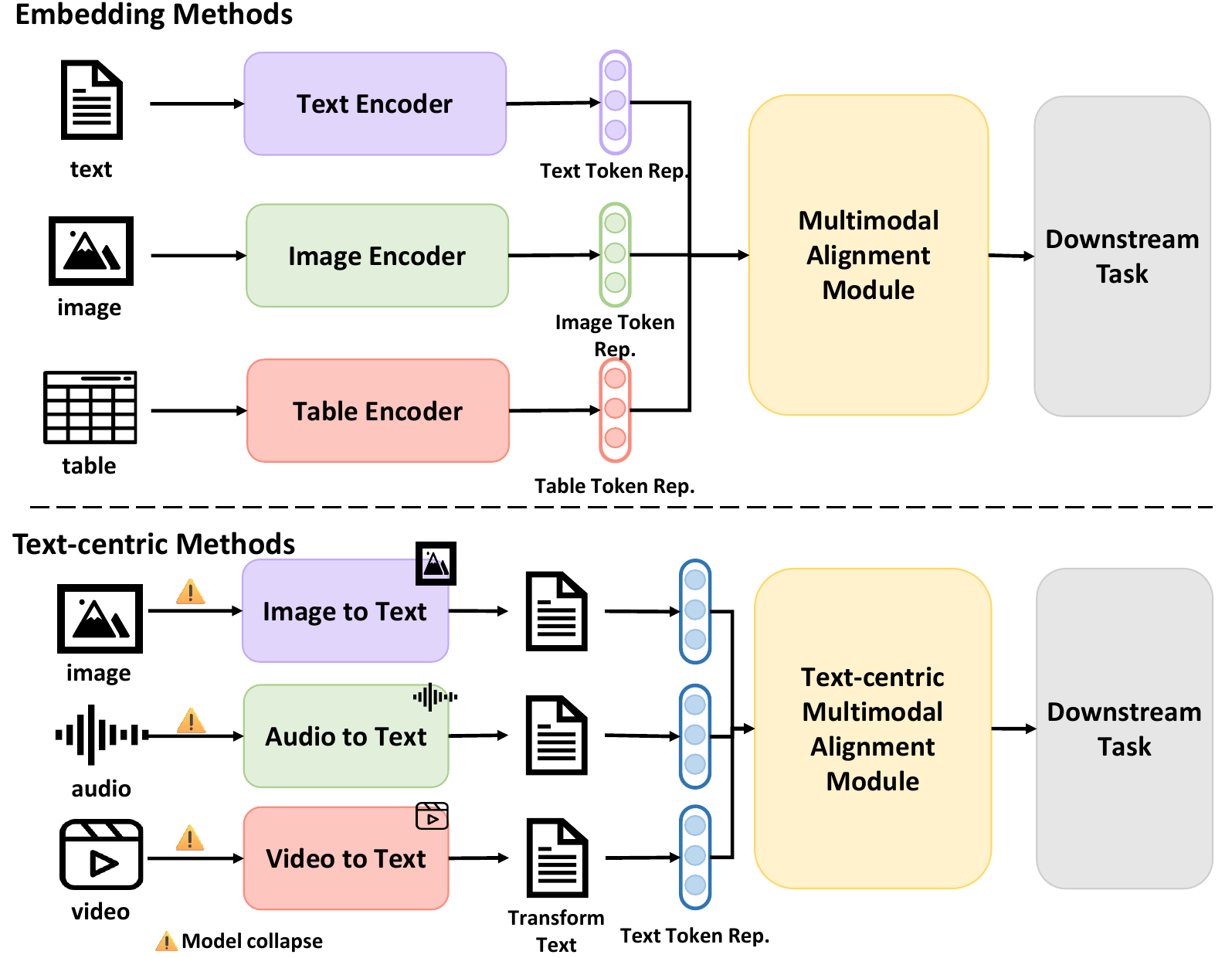}
    \caption{
    Text-centric multimodal alignment, which converts different modalities into text to serve as input prompts for LLMs, is a common method for aligning large multimodal language models when pairwise multimodal data is limited. The potential model collapse phenomenon can jeopardize the robustness of the aligned representation.
    }
    \label{fig:embedding_based}
    \vspace{-0.5cm}
\end{figure}

Recent study~\cite{wang2023too} discovered that Vision LLMs trained on pure synthetically generated high-quality captions by image caption models to replace original noisy data fall into model collapse~\cite{robinson2021can}. This phenomenon can be explained by captioning collapse~\cite{vinyals2015show,wang2020overview} and the one-to-many problem~\cite{young2014image} in image captioning. That is, when transforming images into text, it generates fixed or similar captions for different images, which limits diversity in the output and could potentially jeopardize the downstream model training. This could cause the learned multimodal representations to be less robust for discriminative models (ex. classifier) and cause modality collapse issue for generative models (ex. MLLMs). This leads to the concern where text-centric alignment method would be lead to less robust performance.

In this paper, we improve the modality robustness in text-centric modality alignment methods. Specifically, we aim to repair the modal collapse issue when transforming various modalities into text leads to the generation of fixed or similar outputs, resulting in information loss and reduced diversity. This, in turn, compromises the robustness of the learned multimodal representation.
We further propose using adversarial prompting~\cite{yang2024prompt,dong2023promptattack,xu2024linkprompt} and formulate a text-centric adversarial training approach to enhance the modality robustness of text-centric multimodal alignment. Before converting different input modalities into text using expert models and align these modalities within a similar semantic space, we applied a LLM-based perturbation module on top and increase the diversity and robustness of text representations. This adversarial training procedure along with multimodal alignment will optimizes for a more robust performance. This can be easily understand as using LLMs as an adversary to force improve the robustness of multimodal alignment and the downstream model.

In our experiment, different input modalities are converted into text descriptions using expert foundation models for each modality.
To evaluate the robustness of these representations, we follow the MULTIBENCH~\cite{liang2021multibench} framework, which introduces varying levels of and imperfections. This approach simulates real-world conditions, allowing us to assess how well our unified textual representations perform under scenarios of missing or noisy data.
By rigorously testing under these conditions, we demonstrate that our enhancement can significantly improve the modality robustness. Qualitative analysis also shows that modality summarization and reasoning augmentation with LLMs offer significant advantages: 1) recovering dropped or corrupted information, 2) transforming implicit relationships into explicit text descriptions, and 3) compensating missing information using LLMs as external knowledge sources. These enhancements contribute to the overall robustness and utility of the multimodal representations.

Our contributions are summarized as follows:
\begin{itemize}
    \item We are the first to investigate modality robustness in text-centric alignment methods, revealing their inherent lack of robustness.
    \item We propose an text-centric adversarial training to enhance the robustness for text-centric alignment that demonstrates effective enhancement to modality robustness, consistently outperforming the baselines including traditional robust training methods and multimodal foundation models.
    \item We provide a qualitative analysis illustrating how large language models (LLMs) strengthen the robustness of textual representations in multimodal alignment.
\end{itemize}

\begin{figure*}[t!]
\vspace{-1cm}
  \centering
  \includegraphics[width=0.95\linewidth]{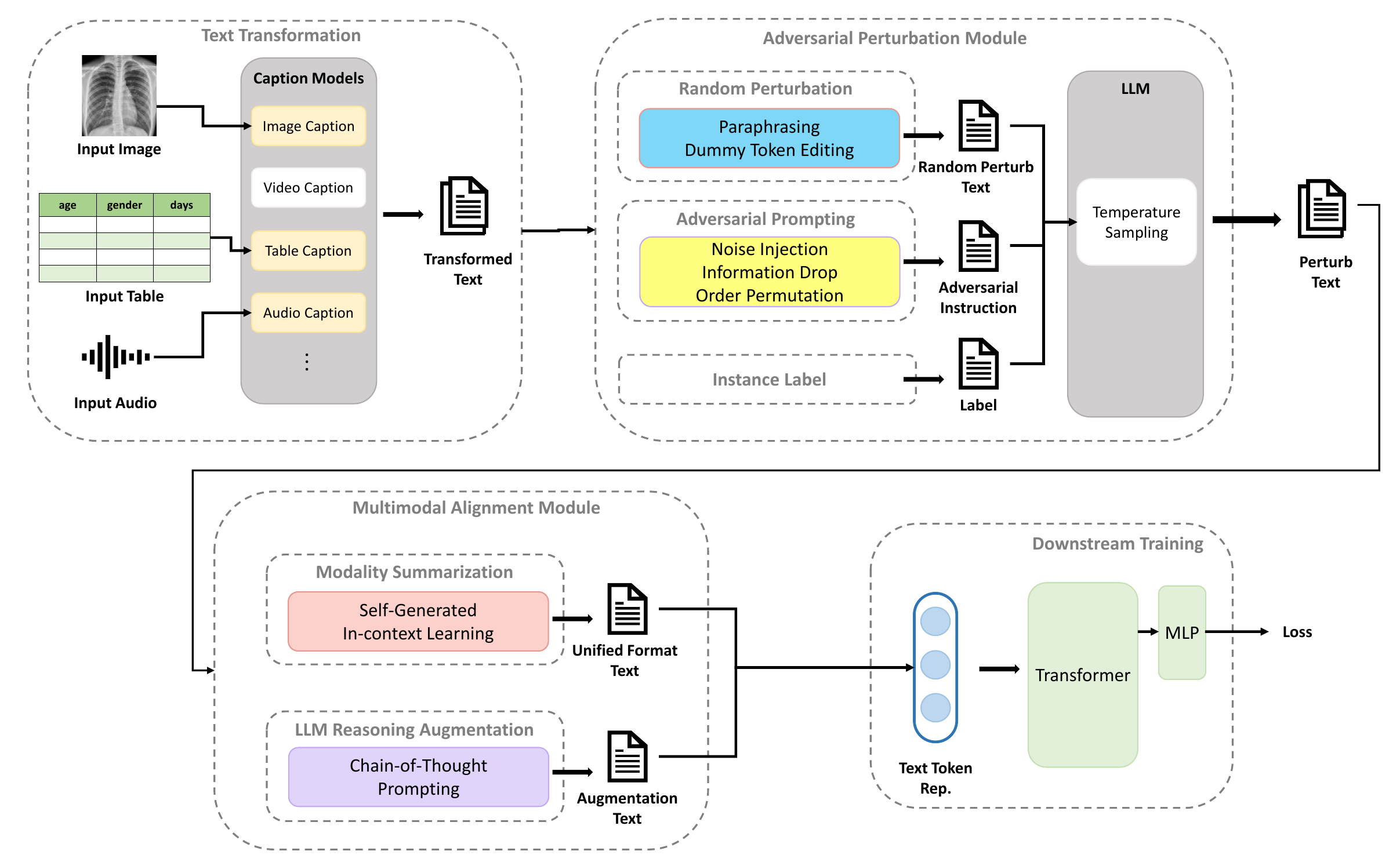}
  \caption{Each raw input modality is transformed into text representations using a corresponding foundation model. Following modality summarization and LLM reasoning are applied in parallel. Finally, the output texts are concatenated as the input to a transformer model for downstream prediction. The inference phase follows a similar pattern. We apply a one-shot in-context learning approach to adapt the linguistic style as anticipated during training.}
  \label{fig:method}
\end{figure*}

\nocite{tsai2021toward}

\section{Related Work}
\label{sec:related}

\subsection{Text-centric Multimodal Alignment}
In recent advancements, several studies have demonstrated the effectiveness of text-centric alignment. For instance, LLaVA~\cite{liu2023llava} utilizes GPT-4 to generate captions and textual descriptions from images, while VideoChat-Text~\cite{li2023videochat} encodes video content into textual formats. In the medical domain, models like OphGLM~\cite{gao2023ophglm} and ChatCAD~\cite{wang2023chatcad} extract information from medical images and convert it into diagnostic reports, seamlessly integrating visual data with textual inputs for LLMs. TAMML~\cite{tsai2024text} converts different input modalities into text for downstream model training and demonstrates significant improvements in handling unseen and diverse modality at test time. These approach depends on the quality of transformed text but offers a straightforward way to achieve multimodal integration.

\subsection{Robustness in Multimodal Learning}
Modality robustness~\cite{ma2022multimodal} addresses the issue of different modalities displaying various noise typologies and the potential for real-world multimodal signals to suffer from missing or noisy data in at least one of the modalities. Similar challenges have been identified in text-centric multimodal alignment methods. Wang et al.~\cite{wang2023too} discovered that Vision LLMs trained on purely synthetically generated high-quality captions by image caption models, intended to replace original noisy data, suffer from model collapse~\cite{robinson2021can}. This phenomenon can be attributed to captioning collapse~\cite{vinyals2015show,wang2020overview} and the one-to-many problem~\cite{young2014image} in image captioning. When transforming images into text, these models generate fixed or similar captions for different images, limiting diversity in the output and leading to trivial solutions.

\subsection{Adversarial Prompting}

Adversarial prompting exposes vulnerabilities in large language models (LLMs) by manipulating their outputs through various techniques. One such technique, \textit{prompt injection}~\cite{liu2023prompt}, involves embedding malicious instructions within prompts to alter the intended response of the model, potentially generating harmful or inappropriate content. Another significant method is \textit{prompt leaking}~\cite{perez2022ignore,hui2024pleak}, where crafted prompts extract sensitive information embedded within the model’s responses, compromising confidentiality. \textit{Jailbreaking}~\cite{ma2024jailbreaking,chao2023jailbreaking,liu2023jailbreaking} techniques bypass the safety mechanisms of LLMs, enabling the model to produce outputs that violate its ethical guidelines.

Additionally, adversarial prompting has been employed to generate adversarial examples. Techniques such as the Prompt-based Attack Approach (PAT)~\cite{yang2024prompt,dong2023promptattack,xu2024linkprompt} generate adversarial examples via mask-and-filling, exploiting the robustness defects of LLMs. These methods have demonstrated high attack success rates, producing diverse, fluent, and natural adversarial examples that can used to significantly improve the robustness of NLP models.

\section{Robust Text-centric Multimodal Alignment}
\label{sec:method}




\begin{figure*}[t]
\vspace{-1.5cm}
    \centering
    \includegraphics[width=1.0\linewidth]{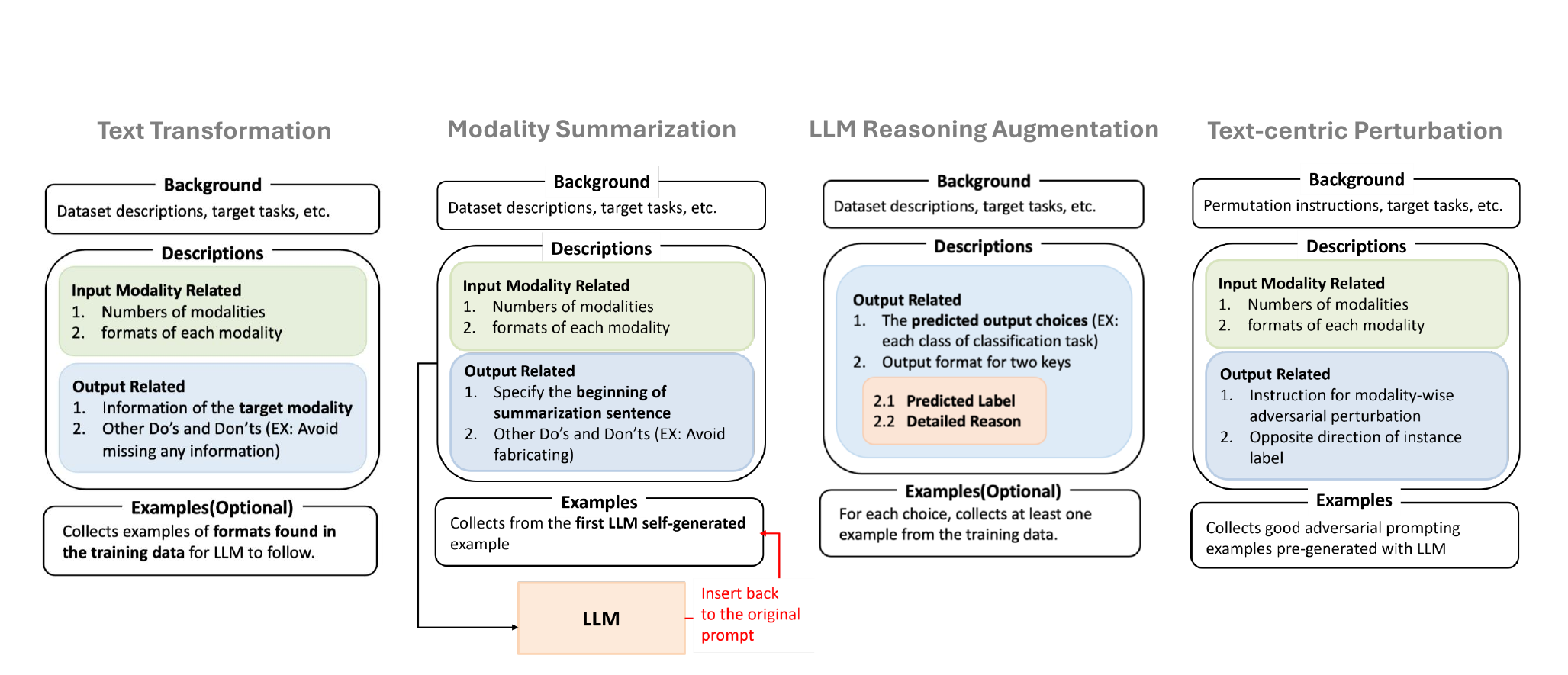}
    \caption{Examples of prompt templates for each module and the required information for input output specified.}
    \label{fig:prompts}
\end{figure*}

This section discusses how we convert raw inputs from different modalities (e.g., images, tabular data) into text representations and apply adversarial prompting to improve the model's robustness. Section~\ref{sec:method:alignment} introduce a text-centic multimodal alignment module. It convert each modality's input into a text representation and align each input modalities. Section~\ref{sec:method:perturb} introduce a perturbation module designed for improving modality robustness by adversarial prompting. The entire process is illustrated in Figure~\ref{fig:method}.

\subsection{Text-centic Multimodal Alignment Module}
\label{sec:method:alignment}
The multimodal alignment module employs LLMs for data transformation across various modalities, aiming to create a unified semantic space. This process is conducted exclusively through in-context learning. Initially, we transform different modalities into text using specialized expert models. Following this, we conduct modality summarization and engage LLMs in text-style translation across modalities. This ensures that all textual representations adopt a consistent linguistic structure, reducing the gap between different modalities and aligning them within a closer semantic space. This step also removes redundant information and mitigates the heterogeneity inherent in text data from diverse sources. Lastly, we include a reasoning augmentation step akin to the Chain-of-Thought method~\cite{wei2022chain}, enhancing the data with LLMs to boost prediction and judgment capabilities. Moreover, we leverage LLMs as a source of large-scale external knowledge, enriching data understanding and interpretative depth~\cite{KDA_2023}.

\subsubsection{Text Transformation}
\label{sec:method:transformation}
In this study, we introduce unique text-based representations for various modalities, enhancing model robustness through rule-based methods and a pre-trained MLLM model. We convert raw inputs into a standardized text format, minimizing the need for modality-specific adaptations and reducing data complexity. This method captures vital information while filtering out noise, boosting the model's ability to handle diverse modalities.

For image data, we use a SOTA image captioning model to produce detailed textual descriptions, converting visual content into text. Textual data remains in its original form to preserve linguistic integrity. For tabular data, we apply a simple serialization method from TabLLM \cite{hegselmann2023tabllm}, structured as "The column name is values," proven to surpass zero-shot learning in LLMs. The transformed texts from each modality are then merged and used as input for further processing.

\subsubsection{Modality Summarization}
\label{sec:method:summarization}
Although all types of data are converted into textual representation, there are still syntactic and semantic gaps between the transformed text across different modalities. In this step, we extend similar linguistic styles text representations to all modalities , improving information quality by facilitating interactions that generate new insights, emphasize key shared data, and remove redundancies. The summary of these modalities is produced by LLMs. Our methodology involves two phases: initially, we collect samples using predefined linguistic styles in prompts that guide the LLMs to merge information from various modalities into a concise summary. Subsequently, this output is integrated with our original prompts, forming a demonstration for in-context learning applied to subsequent samples that follow to the style established in the initial phase.

\subsubsection{LLM Reasoning and Augmentation}
\label{sec:method:reasoning}
We employ LLMs for reasoning based on the Chain-of-Thought method~\cite{wei2022chain} and make LLMs as a large-scale external knowledge source similar to~\cite{KDA_2023} for data augmentation. By assigning prediction tasks with clear instructions and examples, LLMs analyze and augment the textual inputs based on its external knowledge. The models generate predictions and detailed explanations for each input, enhancing the data through this predictive reasoning process. A detailed prompt example is shown in Figure~\ref{fig:prompts}.

\begin{figure*}[t]
\vspace{-1cm}
  \centering
  \includegraphics[width=0.31\linewidth]{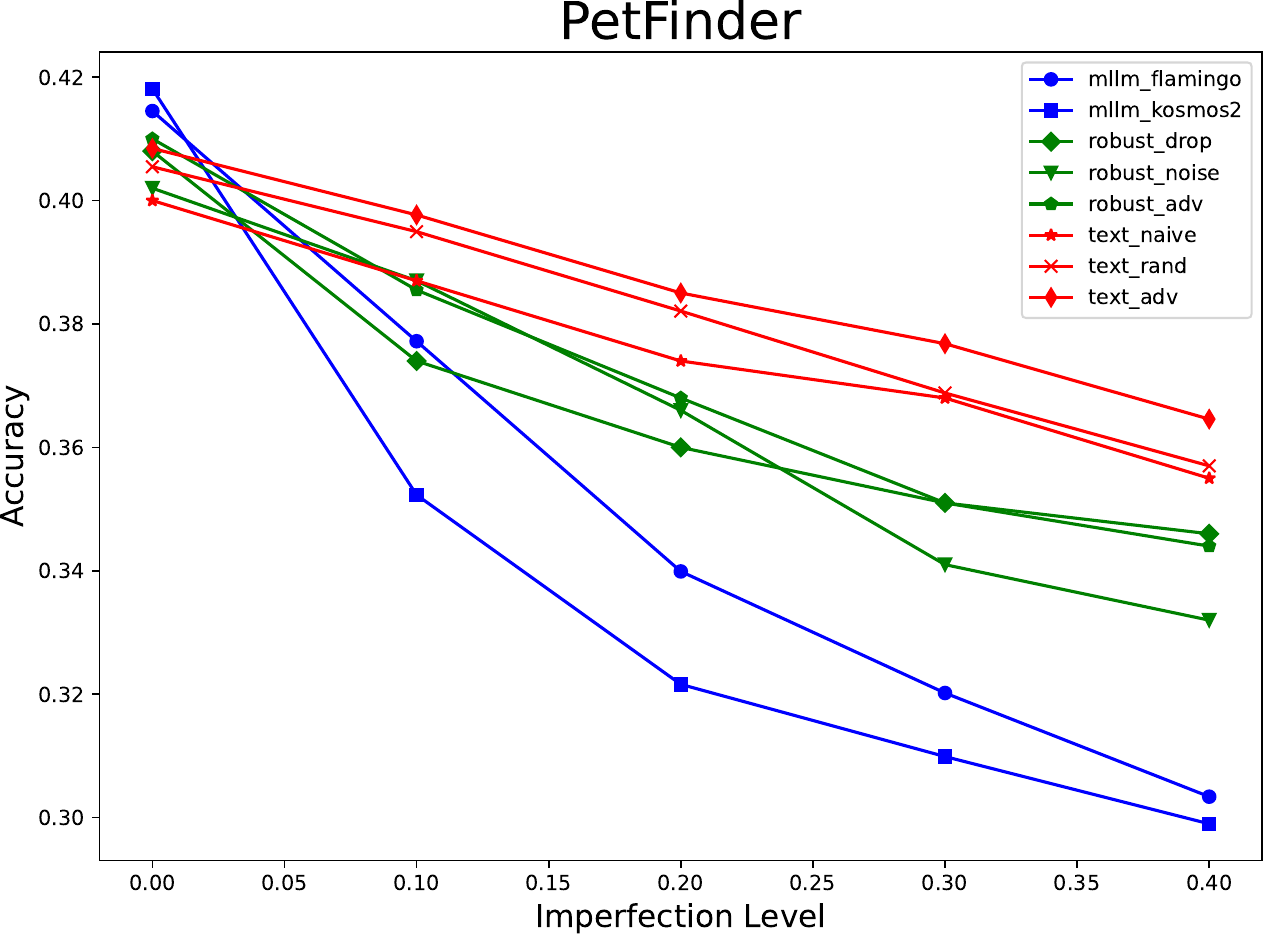}
  \includegraphics[width=0.31\linewidth]{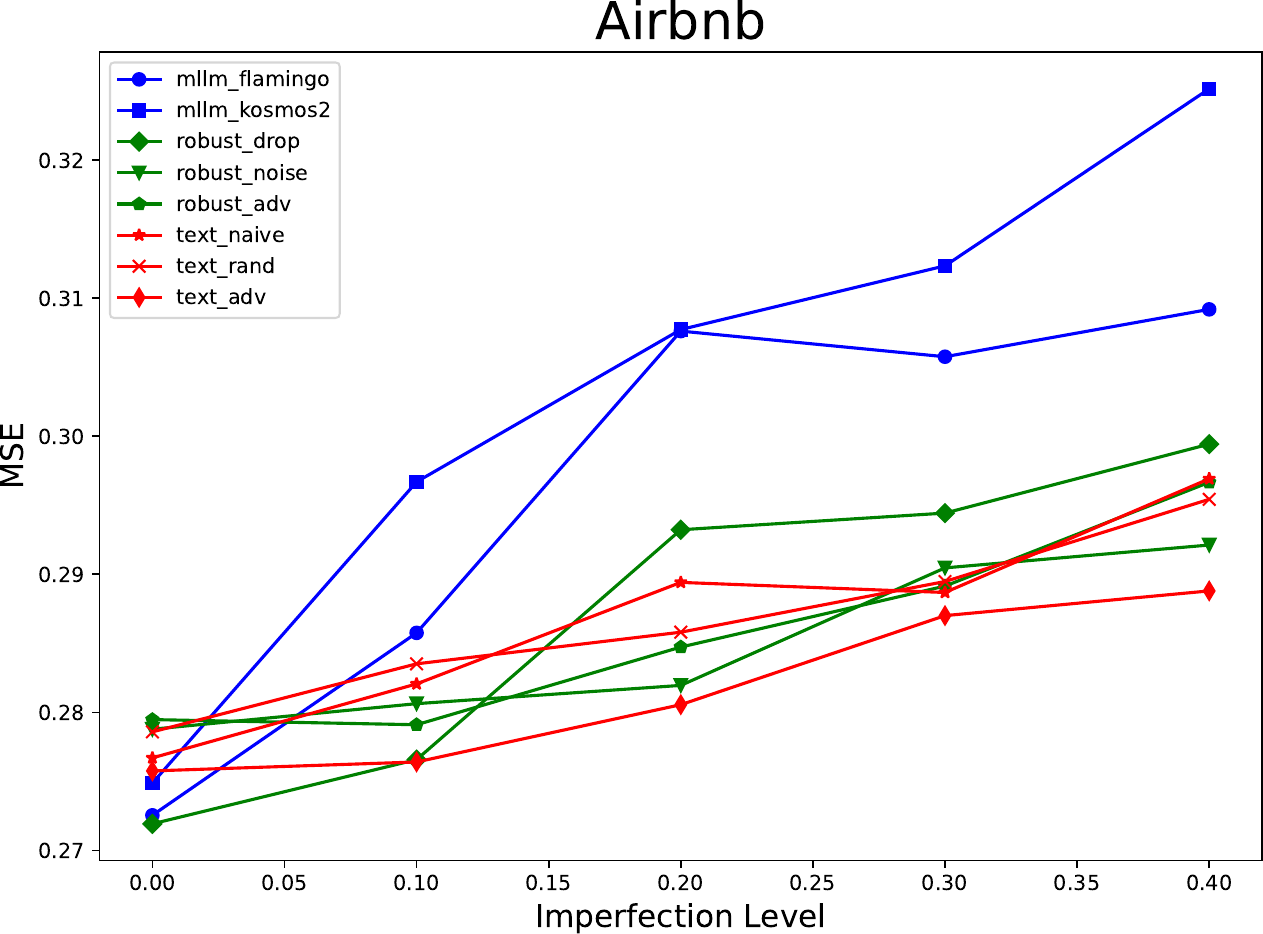}
  \includegraphics[width=0.31\linewidth]{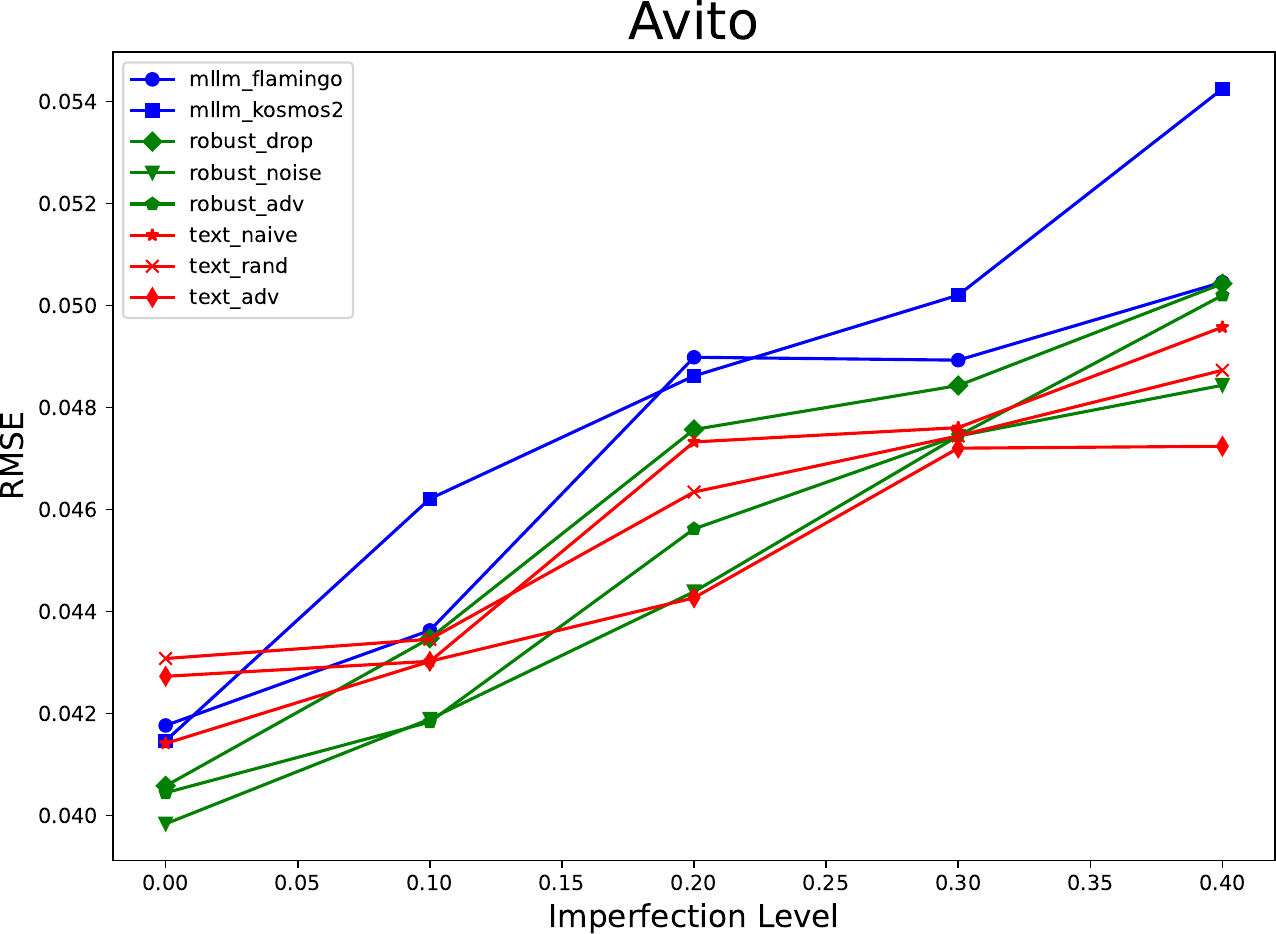}
  \hfill
  \includegraphics[width=0.31\linewidth]{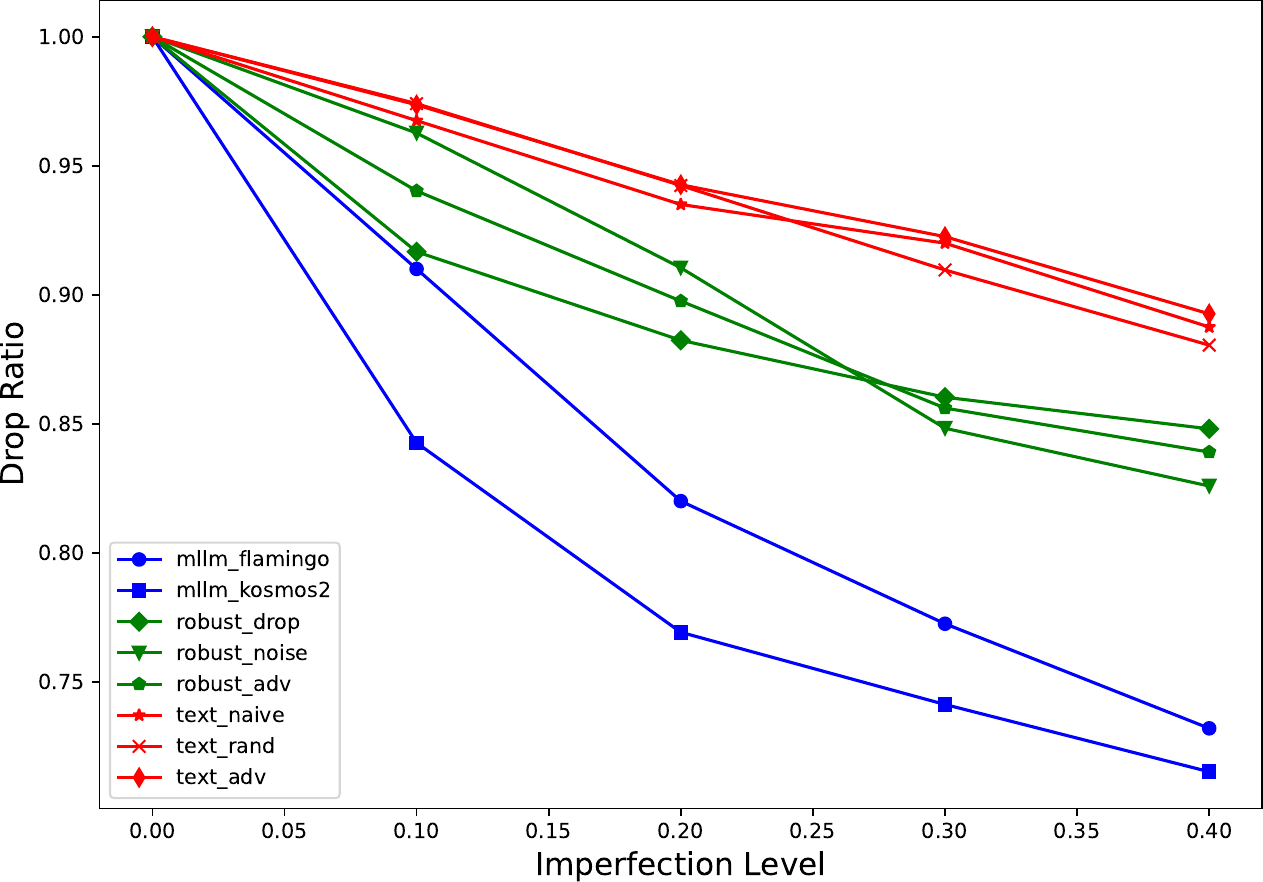}
  \includegraphics[width=0.31\linewidth]{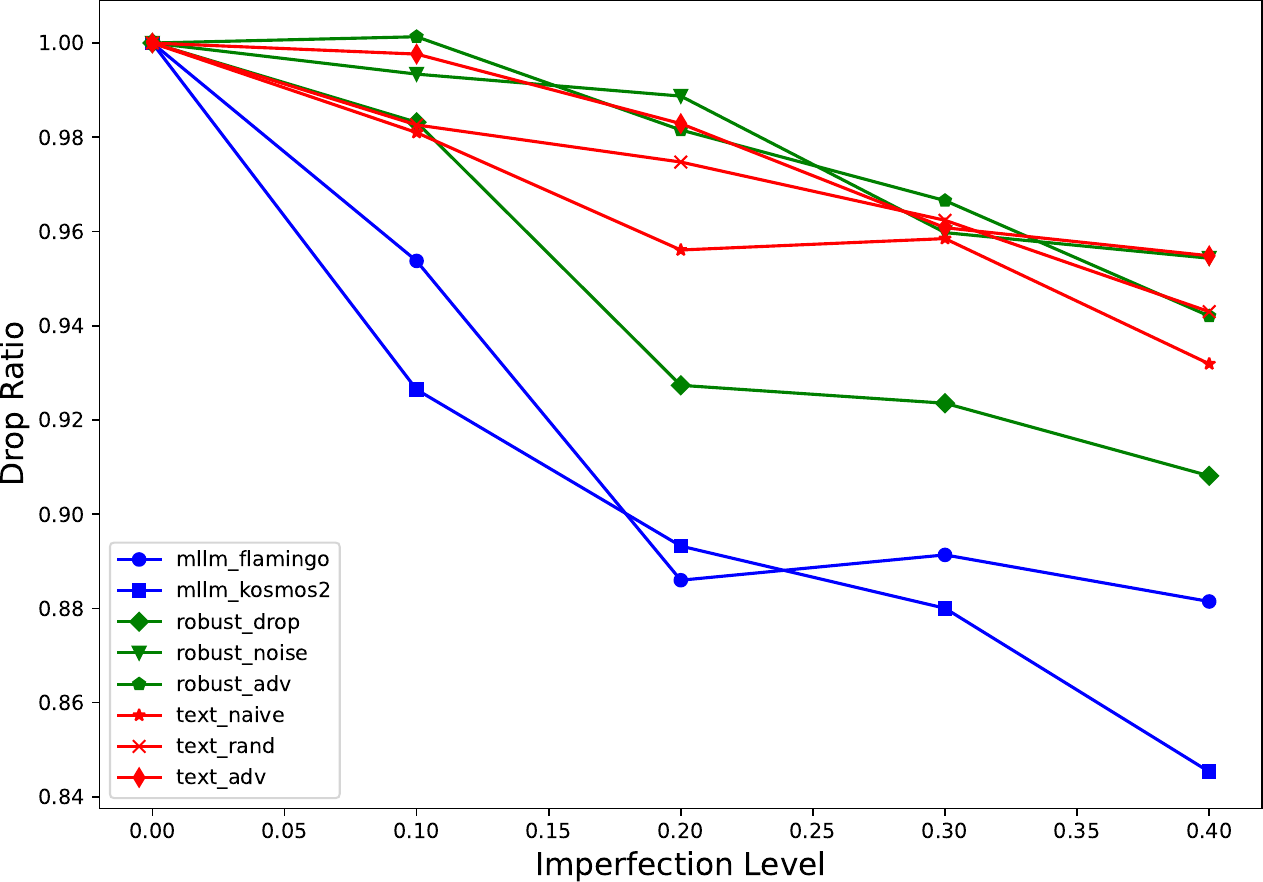}
  \includegraphics[width=0.31\linewidth]{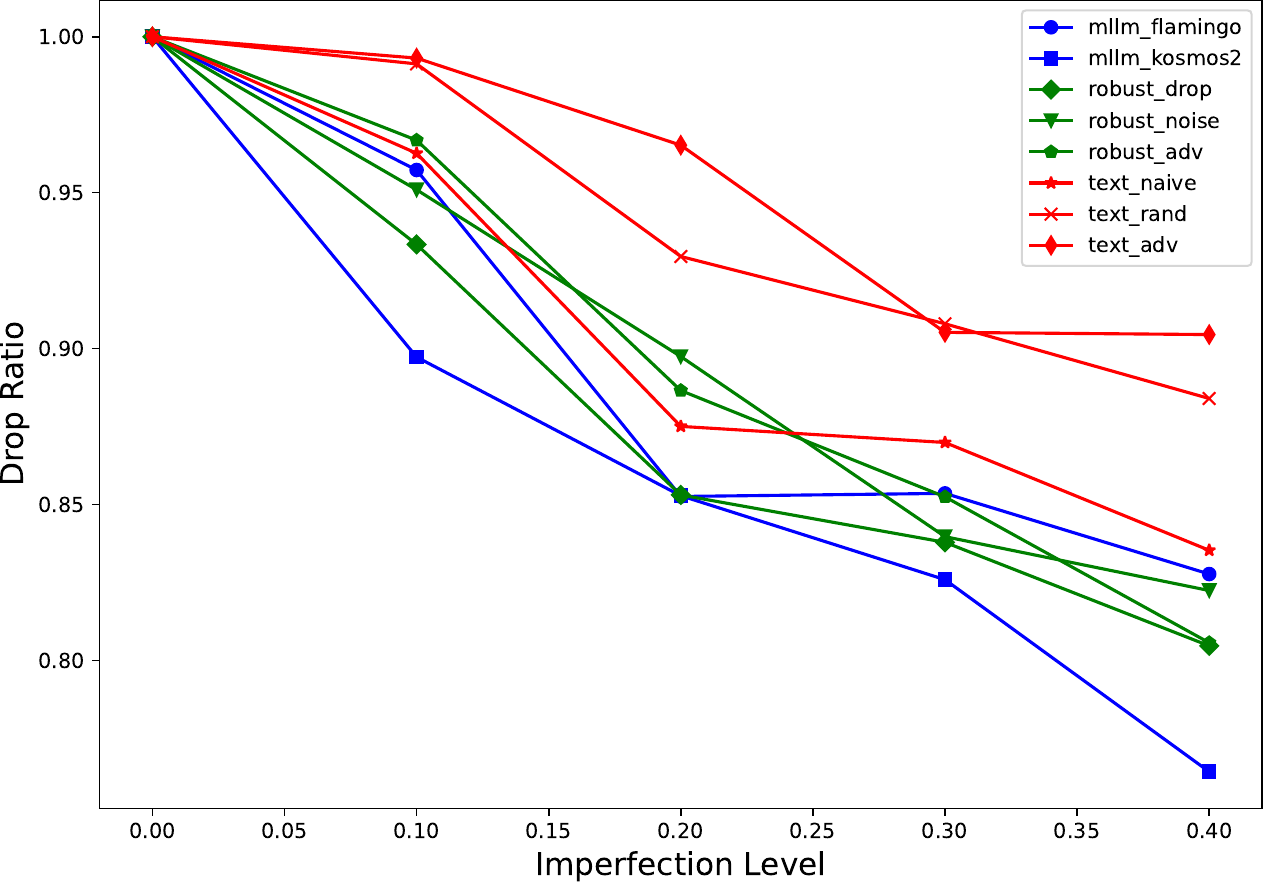}
  \caption{To evaluate the robustness of our model under noisy conditions, we evaluated both relative robustness (top) and effective robustness (bottom) for three datasets. The results from these metrics consistently demonstrate that the text-centric method exhibits superior robustness and resilience to noise when compared to other baseline methods, particularly as noise levels increase. The evaluation was conducted using three different metrics: accuracy, MSE, and RMSE, tailored to each respective dataset.}
  \label{fig:acc_drop_all}
\end{figure*}

\subsection{Text-centric Perturbation Module}
\label{sec:method:perturb}
Perturbation module aims at generating more natural, diverse and fluent adversarial examples in order to overcome the model collapse issue and increase robustness.
In this section, we introduce two types of perturbation module \textit{Random Perturbation} and \textit{Adversarial Perturbation} that both operates on text using LLMs. Random Perturbation
is a naive rule-based prompt perturbation as baseline that randomly derives perturbed variants with Paraphrasing and dummy tokens. Adversarial perturbation employs a set of instruction prompt to guide LLMs in generating adversarial prompts that create the most disruptive examples for various robustness scenarios, effectively shifting the semantics toward the opposite label direction. Both are combined with temperature sampling.

\subsubsection{Random Perturbation}
To generate varied inputs, we employ a paraphrasing technique that queries a language model to produce $k$ paraphrased versions of a given input $x_0$. The process is initiated by using the following prompt: \textit{"Suggest $k$ ways to paraphrase the text above. Remain same semantic}. This approach efficiently generates $k$ distinct paraphrased inputs $\{x_i\}$, for $i = 1, \dots, k$, from a single LLM call, allowing for diverse representations of the original input to be explored.
In addition to paraphrasing, we introduce randomness by edit operations including deletion, insertion and substitution with \textit{dummy tokens} to the original input $x_0$. These tokens, denoted by $d_i$, are selected such that they minimally influence the semantic content of the input. Examples of such tokens include newline characters, tab spaces, ellipses, or additional punctuation marks like extra question marks. The modified inputs are represented as $x_i = x_0 + d_i$ or $x_i = d_i + x_0$, where the random perturbations aim to test the model's sensitivity to minor, non-semantic noise.


\subsubsection{Adversarial Perturbation}

We designed our adversarial perturbation to simulate various scenarios that challenge the modality robustness of text-centric alignment models. Moreover, the generated examples will intentionally manipulate the input modalities and mislead the model’s prediction, thereby creating the most challenging test cases. The process involves crafting specific instruction prompts that guide the language model (LLM) to introduce different types of perturbations into the input data. These perturbations can include noise injection, information dropout, and order permutation, each of which is intended to disrupt the model’s understanding and push its predictions toward incorrect labels.

Our approach begins with the pre-generation of a diverse set of instruction prompts specifically tailored for adversarial purposes. These prompts are crafted to induce the LLM to generate adversarial examples that effectively simulate challenging and unpredictable scenarios. The adversarial perturbations are applied modality-wise, allowing us to evaluate and enhance the robustness of the model across different modalities.
To systematically generate adversarial examples, we follow these steps:

\begin{enumerate}
    \item \textbf{Random Initialization}: We begin by applying random perturbation to the original input $x$ to create an initial variation $x^\prime$. This step ensures that the base input is already altered before applying further adversarial instructions, increasing the likelihood of generating a significantly misleading example.
    
    \item \textbf{Instruction Selection and Parameterization}: An instruction prompt $inst$ is selected from our pre-generated set, which may direct the LLM to perform tasks such as adding noise, dropping critical information, or permuting the order of elements within the input. Alongside the instruction, we set the temperature parameter $T$ to control the randomness of the LLM’s output. A higher temperature may result in more creative and diverse adversarial examples, while a lower temperature produces more deterministic outcomes.
    
    \item \textbf{Adversarial Example Generation}: The LLM generates the adversarial example $x_{adv}$ by completing the instructed operation in a way that most strongly shifts the semantic content towards the opposite label direction. The result is an adversarial example that challenges the model’s ability to maintain accuracy under perturbations. express the process as a formula:
    \[
    x_{\text{adv}} = \text{LLM}(x^\prime, \text{inst}, \text{label}, T)
    \]
\end{enumerate}

This method systematically create samples that simulate real-world scenarios where input modality may be corrupted or misleading. Additionally, this approach supports the iterative refinement of adversarial perturbation.

\section{Experiment}
\label{sec:exp}

We conduct experiments on three multimodal dataset and compared baselines including MLLMs, robust training technique and text-centric approaches. We evaluate the robustness under three different scenarios.
In all experiments, we use Mixtral 8x7B as default language model and GPT-4-Vision for image captioning, unless specified otherwise. For additional results involving different language models, please refer to Table~\ref{tab:ablation:llm}. Furthermore, all trials are run three times and the average is reported. Adversarial training will have maximum ten times more training iterations than regular training.

\subsection{Dataset}
\label{sec:exp:dataset}

\textit{PetFinder.my Adoption Prediction}~\cite{petfinder-adoption-prediction} examines what factors predict how quickly a pet is adopted after being listed. The dataset is a composite of the following modalities:  
\begin{itemize}
    \item Text: contains the description of the status of the pet
    \item Image: contains a profile photo of the pet
    \item Tabular: contains basic information, such as gender and breed. 
\end{itemize}

\noindent\textit{Airbnb Pricing Prediction}~\cite{insideairbnb} is composed of the following modalities used for making a regression prediction of housing prices:
\begin{itemize}
    \item Text: contains the human-written description of the homestay, the neighborhood description, and the host's profile.
    \item Image: contains images of the homestay
    \item Tabular: delivers essential details such as location, rating score, and review counts.
\end{itemize}

\noindent\textit{Avito Demand Prediction}~\cite{avito-demand-prediction} predicts the likelihood of an ad selling something based on user item and context features:
\begin{itemize}
    \item Text: contains the ad title and description.
    \item Image: contains a profile photo of the item.
    \item Tabular: contains basic information, such as region, city, item category, etc.
\end{itemize}

\subsection{Baselines}
\begin{itemize}
    \item \textbf{MLLMs:} We selected two state-of-the-art (SOTA) open-source Multimodal Language Models (MLLMs) for robustness comparison: Kosmos-2~\cite{peng2023kosmos} and Flamingo~\cite{alayrac2022flamingo}. This help show the comparison between large foundation models without robust training.
    
    \item \textbf{Robust Training:} To evaluate the robustness of our text-centric approach against traditional methods, we employed several robust training techniques for the downstream models. These included gaussian noise injection, dropout, and adversarial training using Projected Gradient Descent (PGD)~\cite{madry2017towards}. These baselines help demonstrate whether our text-based method, which leverages LLMs, offers superior robustness compared to traditional embedding-based methods.
    
    \item \textbf{Text-Centric Approaches:} We compared the effects of naive(transform to text with no perturbation), random perturbation and adversarial perturbation to determine whether adversarial prompting provides greater robustness than merely increasing input diversity and text transformation.
\end{itemize}

\subsection{Evaluation}
\label{sec:exp:eval}

\subtwosection{Evaluation Protocol}
To evaluate the robustness of our models, we adopted similar methodologies outlined in MULTIBENCH~\cite{liang2021multibench}. We define the following three scenarios:
\begin{enumerate}
    \item Noisy Modality: For images, we introduced Gaussian noise at five different levels from 10\% to 90\%. For text descriptions, we randomly dropped words with five levels of probability from 10\% to 50\%. For table data, we randomly dropped column features with probabilities from 10\% to 90\%.
    \item Dynamic Modality: Dynamically permute the order of input modalities to test robustness. Text-centric alignment and token-based transformer models should exhibit invariance to the order of tokens within a prompt.
    \item Missing Modality: Randomly select modalities that would be absent at test time. Zero vectors are filled in for robust training.
\end{enumerate}

\subtwosection{Evaluation Metric}
Following our evaluation protocols designed to mimic the modality-specific and multimodal imperfections described in MULTIBENCH, we evaluate both \textit{Accuracy}, \textit{MSE}, \textit{RMSE} under imperfections (relative robustness) and the \textit{Drop ratio} of performance when imperfections are introduced (effective robustness).


\subsection{Noisy Modality Results}
Figure~\ref{fig:acc_drop_all} illustrates that our method consistently achieves the lowest drop ratio under noisy modality conditions, outperforming other baselines, particularly at the highest noise levels. For the Petfinder dataset, our text-centric adversarial method experienced only an 11.3\% drop, significantly outperforming the robust training method at 15.2\% and the MLLMs, which saw a substantial drop of 28.5\%. Similar patterns are observed in the Airbnb and Avito datasets, where our method consistently surpasses all baselines. Additionally, Appendix~\ref{app:sec:extra} Figure~\ref{fig:acc_drop_image}, which applies modality-specific noise across different modalities, reveals that the impact of added noise varies across modalities. This finding opens a future research direction to explore the text-centric modality collapse problem.

\nocite{tsai2023differential}

\begin{table}[t!]
\setlength\tabcolsep{2pt}
\begin{tabular}{|ll|rr|rr|rr|}
\toprule
                    &             & \multicolumn{2}{c|}{Petfinder} & \multicolumn{2}{c|}{Airbnb}           & \multicolumn{2}{c|}{Avito}    \\
\midrule
                    &             & ACC$\uparrow$ & Drop & MSE$\downarrow$   & Drop   & RMSE$\downarrow$ & Drop \\
\midrule
MLLM                & Kosmos2     & .371       & .883 & .285  & .954   & .043 & .953 \\
                    & Flamingo    & .374       & .890 & .283  & .961   & .044 & .931 \\
\midrule
Robust              & Noise       & .296       & .704 & .478  & .569  & .080 & .512 \\
Training            & Dropout     & .313       & .745 & .430  & .632  & .067 & .611 \\
                    & Adversarial & .302       & .719 & .470  & .578  & .069 & .594 \\
\midrule
Text                & Naive       & .386       & .919 & .277  & .981 & .042 & .976 \\
centric             & Random      & .390       & .928 & .280  & .871 & .043 & .953 \\
                    & Adversarial & \textbf{.397}       & \textbf{.945} & \textbf{.274}  & \textbf{.992} & \textbf{.042} & \textbf{.977} \\
\bottomrule
\end{tabular}

\caption{Dynamic Modality Evaluation. Both relative robustness (left) and effective robustness (right) for three datasets are shown. Text-centric adversarial prompting methods outperforms all baselines and show strong invariance to dynamic input order. Robust training technique completely failed as expected.}
\label{tab:dynamic}
\end{table}

\begin{table}[t!]
\setlength\tabcolsep{2pt}
\begin{tabular}{|ll|rr|rr|rr|}
\toprule
                    &             & \multicolumn{2}{c|}{Petfinder} & \multicolumn{2}{c|}{Airbnb}           & \multicolumn{2}{c|}{Avito}    \\
\midrule
                    &             & ACC$\uparrow$ & Drop & MSE$\downarrow$   & Drop   & RMSE$\downarrow$ & Drop \\
\midrule
MLLM                & Kosmos2     & .302       & .719 & .320  & .851   & .050 & .824 \\
                    & Flamingo    & .310       & .738 & .318  & .855   & .051 & .803 \\
\midrule
Robust              & Noise       & .323       & .769 & .319  & .852  & .049 & .836 \\
Training            & Dropout     & .310       & .738 & .320  & .850  & .050 & .824 \\
                    & Adversarial & .330       & .785 & .308  & .883  & .048 & .854 \\
\midrule
Text                & Naive       & .362       & .861 & .309  & .880 & .047 & .872 \\
centric             & Random      & .370       & .881 & .310  & .877 & .047 & .871 \\
                    & Adversarial & \textbf{.378}       & \textbf{.900} & \textbf{.302}  & \textbf{.899} & \textbf{.046} & \textbf{.891} \\
\bottomrule
\end{tabular}

\caption{Missing Modality Evaluation. Both relative robustness (left) and effective robustness (right) for three datasets are shown. Text-centric adversarial prompting methods outperforms all baselines with a large margin.}
\label{tab:missing}
\end{table}

\nocite{tsai2024toward}

\subsection{Dynamic Modality Results}
To evaluate the model's invariance and robustness to different modality input orders, we tested and averaged the results across all possible input permutations. Table~\ref{tab:dynamic} shows that our method has the lowest drop ratio, outperforming all other baselines. For the Petfinder dataset, our text-centric adversarial method experienced only a 5.5\% drop in performance, compared to 11.1\% for the MLLMs, and far better than robust training methods, which performed close to random guessing, as expected. These trends are consistent in the Airbnb and Avito datasets, where our method consistently outperforms all baselines. The token-based, text-centric approach naturally provides an advantage in maintaining robustness against dynamic input orders, underscoring its effectiveness in various scenarios.

\subsection{Missing Modality Results}
To evaluate robustness under conditions of missing modalities at test-time, we tested and averaged the results across different combinations of dropped modalities. Table~\ref{tab:missing} shows that our method achieves the lowest drop ratio, outperforming all other baselines. For the Petfinder dataset, our text-centric adversarial method experienced only a 10\% drop, which is significantly better than the robust training method at 22.5\% and the MLLMs at 26.2\%. Similar observations are made in the Airbnb and Avito datasets, where our method consistently outperforms all baselines, reaffirming its superior robustness in scenarios with missing modalities.

\subsection{Ablation Study}
\label{sec:ablation}

\subsubsection{Module Ablation}
We examine the two primary components of our method: the Alignment Module and the Permutation Module. The results presented in Table~\ref{tab:ablation:llm} indicate that both modules contribute almost equally to the overall performance. When both modules are removed, the performance drops significantly, nearing the levels observed with standard MLLMs.

\subsubsection{Language Model Ablation}
We adopted different LLMs to test whether robustness is consistent across various LLMs and how well it is sustained with different model types and sizes. Table~\ref{tab:ablation:llm} shows that GPT-4o offers the best performance among all LLMs. However, the impact of model type and size is minor, with a maximum difference of around 2\% in accuracy. We conclude that the robustness of our method is transferable between models.

\nocite{tsai2024handling}




\begin{table}[t]
\centering
\begin{tabular}{lccc}
\hline
Model & Noisy & Dynamic & Missing \\
\hline
GPT-4o & \textbf{0.4086} & \textbf{0.398} & \textbf{0.381} \\
GPT-3.5-turbo & 0.4037 & \textbf{0.398} & 0.379 \\
Mixtral8x7b & 0.4033 & 0.397 & 0.378 \\
\hline
w/o alignment & 0.3727 & 0.383 & 0.363\\
w/o perturbation & 0.3659 & 0.386 & 0.362 \\
w/o both & 0.3342 & 0.373 & 0.302 \\
\hline
\end{tabular}
\caption{Ablation study on each module contribution and the impact of different LLMs on PetFinder dataset. Both alignment module and perturbation module is necessary to perform well. GPT-4o offers the best performance, but the impact between LLMs is not substantial and, at max, $\sim$2\% accuracy.}
\label{tab:ablation:llm}
\end{table}

\nocite{tsai2023rtlfixer}

\section{Qualitative Analysis and Findings}
\label{sec:analysis}

In this section, we delve into the qualitative aspects and explore how LLMs can effectively handle scenarios where information is either incomplete or lost across different modalities and how LLMs compensate these lost information.

\begin{figure}[ht!]
\begin{tcolorbox}[width=1.0\linewidth, halign=left, colframe=black, colback=white, boxsep=0.01mm, arc=1.5mm, left=2mm, right=2mm, boxrule=0.5pt]\footnotesize

\textbf{Text with noise}\\ 
\textcolor{gray}{The type of pet is Dog. The name of pet is Filo. \textcolor{red}{The age of pet when listed (in months) is 78.}The color 1 of pet is Brown. The fur length of pet is Short.  The primary breed of pet is Mixed Breed. \textcolor{red}{The secondary breed of pet (if pet is of mixed breed) is Unknown.}
}

\textbf{Summarization}\\ 
\textcolor{blue}{This pet is a gentle and handsome 6-year-old mixed breed dog named Filo. He has a light brown, short coat and soft brown eyes.} 

\end{tcolorbox}

\vspace{-10pt}
\caption{The tabular data has dropped the color and fur length column (gray). However, it was recovered (blue) after applying alignment module with LLM that compensate the information from input image.}
\label{fig:analysis-table}
\end{figure}

\begin{figure}[h!]
\vspace{-10pt}
\begin{tcolorbox}[width=1.0\linewidth, halign=left, colframe=black, colback=white, boxsep=0.01mm, arc=1.5mm, left=2mm, right=2mm, boxrule=0.5pt]\footnotesize

\textbf{Text with noise}\\ 
\textcolor{red}{Jack experienced it Jack dog even is He recovering but special care skin which hair to Vet this was to Hopefully lover who take good dog a}

\textbf{Summarization}\\ 
\textcolor{blue}{Although the profile write-up for Jack Jack is somewhat fragmented, it indicates that he has experienced some challenges but is on the road to recovery} 

\textbf{Reasoning}\\ 
\textcolor{cyan}{3. **Health Condition**: Jack Jack has a minor injury and is recovering. Potential adopters might be hesitant to take on a pet that requires special care, even if the injury is minor.}

\end{tcolorbox}

\vspace{-10pt}
\caption{This example demonstrates that alignment module can compensate for noisy text input, transforming it into clear descriptions.}
\label{fig:analysis-text}
\end{figure}

\subtwosection{LLMs recover lost data from other modalities}
Figure~\ref{fig:analysis-table} illustrates that even when critical information is lost from one modality, our approach effectively leverages data from other available modalities to reconstruct the missing content. This ability highlights the strength of multimodal learning, where the complementary information across different modalities compensates for gaps, ensuring robust data recovery. Detailed input examples and reconstructions are provided in Appendix.

\subtwosection{LLMs compensate missing information with knowledge and transform implicit relations into explicit text description by reasoning.}
In scenarios where input text is fragmented due to word dropout, and no relevant data is available from other modalities, Figure~\ref{fig:analysis-text} demonstrates how our method utilizes the extensive knowledge embedded in LLMs. The model not only reconstructs the missing words but also enhances the coherence of the text by drawing on contextual understanding and reasoning capabilities. This allows the LLM to infer and explicitly articulate underlying meanings that were only implicit in the original input.

\nocite{da2022fast}

\section{Conclusion}
\label{sec:conclusion}
This study evaluates the robustness of text-centric multimodal alignment, revealing that it is less robust compared to other robust training techniques. To address this, we propose a novel adversarial training approach specifically designed for text-centric alignment, which outperforms existing baselines that demonstrates strong resistance to noise, input permutation, and missing modalities. Ablation studies further highlight both multimodal alignment and adversarial permutation modules are crucial for enhancing robustness. Additionally, our method is highly transferable across different LLMs. These insights contribute to the development of more resilient multimodal alignment techniques.

\bibliography{aaai25}

\clearpage

\appendix
\onecolumn

\section{Appendix}
\label{sec:appendix}

In Section~\ref{sec:analysis}, we only highlight the most important parts in qualitativfe analysis. In this section, we will present the detailed raw input for a more comprehensive analysis.

\subsection{Noise in Text Data}
\label{sec:appendix:text}

\begin{figure*}[ht]

\begin{tcolorbox}[width=1.0\linewidth, halign=left, colframe=black, colback=white, boxsep=0.01mm, arc=1.5mm, left=2mm, right=2mm, boxrule=0.5pt]\footnotesize
\textbf{Text transformation - text}\\ 

The type of pet is Dog. The name of pet is Jack Jack. The age of pet when listed (in months) is 36. The gender of pet is Male. The color 1 of pet is Brown. The color 2 of pet is Unknown. The color 3 of pet is Unknown. The size of pet at maturity is Small. The fur length of pet is Short. The vaccination status of pet is Yes. The deworming status of pet is Yes. The sterilisation status of pet is No. The health condition of pet is Minor-Injury. The number of pets represented in profile is 1. The pet adoption fee is Free. The state location of pet in Malaysia is Selangor. The total uploaded videos for this pet is 0. The total uploaded photos for this pet is 3. The primary breed of pet is Chihuahua. The secondary breed of pet (if pet is of mixed breed) is Unknown.

The image presents a small, fawn-colored dog with large ears that stand upright, creating an alert and inquisitive expression. The dog is looking up towards the camera with its head slightly tilted, which may indicate curiosity or attentiveness. It is wearing a light blue collar with a bell, signifying that it is a pet likely kept indoors. The dog is positioned on a plain tiled floor, with the low angle perspective of the shot highlighting its petite stature and endearingly disproportionate ear-to-head size ratio.

\textcolor{red}{The profile write-up for this pet is : Jack experienced it Jack dog even is He recovering but special care skin which hair to Vet this was to Hopefully lover who take good dog a}
\end{tcolorbox}

\begin{tcolorbox}[width=1.0\linewidth, halign=left, colframe=black, colback=white, boxsep=0.01mm, arc=1.5mm, left=2mm, right=2mm, boxrule=0.5pt]\footnotesize

\textbf{After Summarization and Reasoning - text}\\ 
This pet is a charming 36-month-old male Chihuahua named Jack Jack, who has a short brown fur coat. Despite a minor injury, he is in good health and has been both vaccinated and dewormed, although he has not been sterilized. Jack Jack is a small-sized dog who currently resides in Selangor, Malaysia, and is available for adoption free of charge. Jack Jack's profile includes three photos, one of which captures him as a small, fawn-colored dog with large, upright ears that give him an alert and inquisitive look. He is seen wearing a light blue collar with a bell, suggesting he is an indoor pet. The image shows him on a plain tiled floor, with his head slightly tilted up towards the camera, highlighting his petite stature and the endearing size ratio of his ears to his head.
\textcolor{blue}{Although the profile write-up for Jack Jack is somewhat fragmented, it indicates that he has experienced some challenges but is on the road to recovery. He requires special care for his skin and fur, and hopefully, a dog lover will come forward to provide him with the good care he deserves.} For more information or photos, interested adopters can reach out via Whatsapp. Several factors influence the adoption rate for Jack Jack, and while he has some positive attributes, there are also some challenges that may slow down his adoption.

1. **Breed and Size**: Jack Jack is a Chihuahua, a breed that is generally popular due to their small size and suitability for apartment living. This is a positive factor that could help in his adoption.

2. **Age**: At 36 months (3 years old), Jack Jack is no longer a puppy. Puppies tend to get adopted faster than adult dogs, which may slow down his adoption rate.

\textcolor{cyan}{3. **Health Condition**: Jack Jack has a minor injury and is recovering. Potential adopters might be hesitant to take on a pet that requires special care, even if the injury is minor.}

4. **Vaccination and Deworming**: The fact that Jack Jack is vaccinated and dewormed is a positive aspect and can reassure potential adopters about his health.

5. **Sterilisation Status**: Jack Jack is not sterilized, which could be a concern for some adopters. Many adopters prefer pets that are already neutered to avoid the cost and responsibility of the procedure.

6. **Profile Write-Up**: The profile write-up for Jack Jack is somewhat unclear and lacks detailed information that could appeal to potential adopters. A well-written profile can significantly impact adoption rates.

7. **Photos**: There are only 3 photos uploaded. More photos and possibly videos could help showcase Jack Jack's personality and increase interest.

8. **Location**: Jack Jack is located in Selangor, Malaysia, which has a significant population and potentially a larger pool of adopters. However, the adoption rate can still vary based on local preferences and the number of available pets.
\end{tcolorbox}
\caption{LLM summarization utilizes its capabilities to infer the original meaning of sentences obscured by noise. LLM Reasoning further transforms implicit relations into explicit text descriptions. (red: noisy text, blue: results inferred through summarization, light blue: potential meaning behind the information uncovered through reasoning.) This example demonstrates the effectiveness of our text-centric methods.}
\label{fig:appendix-analysis-text}
\end{figure*}

\begin{figure*}
\subsection{Noise in Tabular Data}
\label{sec:appendix:image}
\end{figure*}

\begin{figure*}[ht]

\begin{tcolorbox}[width=1.0\linewidth, halign=left, colframe=black, colback=white, boxsep=0.01mm, arc=1.5mm, left=2mm, right=2mm, boxrule=0.5pt]\footnotesize

\textbf{Text transformation - table}\\ 
\textcolor{gray}{The type of pet is Dog. The name of pet is Filo. \textcolor{red}{The age of pet when listed (in months) is 78.} The gender of pet is Male. The color 1 of pet is Brown. The color 2 of pet is Unknown. The color 3 of pet is Unknown. The size of pet at maturity is Large. The fur length of pet is Short. The vaccination status of pet is Yes. The deworming status of pet is Yes. The sterilisation status of pet is Unsure. The health condition of pet is Healthy.\textcolor{red}{ The number of pets represented in profile is 1.} The pet adoption fee is Free. The state location of pet in Malaysia is Kuala Lumpur. The total uploaded videos for this pet is 0.\textcolor{red}{The total uploaded photos for this pet is 5.} The primary breed of pet is Mixed Breed. \textcolor{red}{The secondary breed of pet (if pet is of mixed breed) is Unknown}}

\textcolor{violet}{The image shows a dog with a light brown, short coat lying comfortably on a blue plastic crate.} The dog has a gentle expression, with soft brown eyes, and appears to be calmly resting inside what looks like a tiled room, likely a part of its home. The dog's collar with an identification tag suggests it is well-loved and cared for. The environment looks clean and provides a sense of a safe and secure space for the pet to relax.

The profile write-up for this pet is : Meet Filo, the cute and handsome 6 year old mixed breed boy. Filo is a lucky stray saved by a kind owner earlier this year. \textcolor{violet}{This kind owner brought Filo to the vet to treat some external wounds.} However, the kind owner has 2 dogs and don't have room for Filo. Filo continues to stay in the cage with the vet from that day till now while his wounds are treated. When the vet is available, Filo gets to go out for a short walk. If the clinic is packed, Filo stays in the cage for the whole day. Its not anybody's fault. Filo is still grateful for this kind owner and the kind vet for their hospitality. But Filo's future is still very uncertain and he is hoping that he's lucky to be granted a permanent loving home. I met Filo when I brought my poodle to visit the clinic. I learn of Filo's story from this kind vet. If you can give Filo a loving home or if you need more information, please call me. 

\end{tcolorbox}

\begin{tcolorbox}[width=1.0\linewidth, halign=left, colframe=black, colback=white, boxsep=0.01mm, arc=1.5mm, left=2mm, right=2mm, boxrule=0.5pt]\footnotesize

\textbf{After Summarization and Reasoning - table}\\ 
\textcolor{blue}{This pet is a gentle and handsome 6-year-old mixed breed dog named Filo. He has a light brown, short coat and soft brown eyes,} giving him a calm and serene expression. Filo's story is one of resilience and hope. He was a stray who was fortunate to be rescued by a kind owner earlier this year. This compassionate person took Filo to the vet to treat some external wounds, but due to having two other dogs, they couldn't keep Filo at home. Since then, Filo has been staying at the vet's clinic, where he continues to receive care and treatment for his wounds.

Filo's life at the clinic is a mix of confinement and occasional freedom; he stays in a cage most of the day but gets to go out for short walks when the vet is available. Despite these limitations, Filo remains grateful for the kindness shown to him by both his rescuer and the vet. However, his future remains uncertain, and he longs for a permanent, loving home where he can feel safe and secure.

Filo's profile features five photos, one of which shows him lying comfortably on a blue plastic crate inside a tiled room, likely part of his temporary home at the clinic. The clean and secure environment, along with his collar and identification tag, suggests that he is well-loved and cared for. If you can provide Filo with the forever home he deserves or need more information, please call the number provided. Filo's adoption rate is likely to be more than 3 months for several reasons:

1. **Age**: At 78 months (6.5 years old), Filo is significantly older than the other pets listed. Younger pets, especially puppies and kittens, generally have higher adoption rates because they are perceived as more adaptable and have a longer potential lifespan.

\textcolor{blue}{2. **Health and Background**: While Filo has been treated for his external wounds, his background as a stray and his extended stay in a cage might raise concerns for potential adopters about his health and behavior. The write-up mentions his gratefulness and calm demeanor, but it does not provide detailed information about his current health status or behavior, which could be crucial for potential adopters.}

3. **Emotional Connection**: The profile write-up is heartfelt and provides a touching backstory, but it lacks the emotional appeal seen in the profiles of younger pets. The language used to describe Filo's situation is more factual and less engaging compared to descriptions of other pets, which emphasize their cuteness and playful nature.

4. **Visual Appeal**: Although there are 5 photos, the description of the image shows Filo in a resting position, which may not be as engaging as images of playful or interactive behavior. Potential adopters often respond more positively to images that show the pet's personality and energy.

5. **Competition**: Filo is competing with younger, more visually appealing pets that are often adopted faster. His profile needs to stand out more to attract potential adopters who are specifically looking for an older, more mature dog.

6. **Location and Accessibility**: The profile does not specify the exact location beyond being at a vet clinic, which might make it less accessible for potential adopters who prefer to know more about where the pet is currently staying.
\end{tcolorbox}

\caption{An example of LLM summarization recovering noisy tabular data through other modality information.(gray: dropped columns, red: retained column, purple: data referenced by the summarization, blue: data generated after compensation missing tabular columns information from image.) In this example, even after dropping table information such as the pet's health condition, fur length, and color, these details can still be recovered from images and text descriptions.}
\label{fig:appendix-analysis-table}
\end{figure*}

\clearpage
\subsection{Extra Figures}
\label{app:sec:extra}

\begin{figure*}[h]
  \centering
  \includegraphics[width=1.0\linewidth]{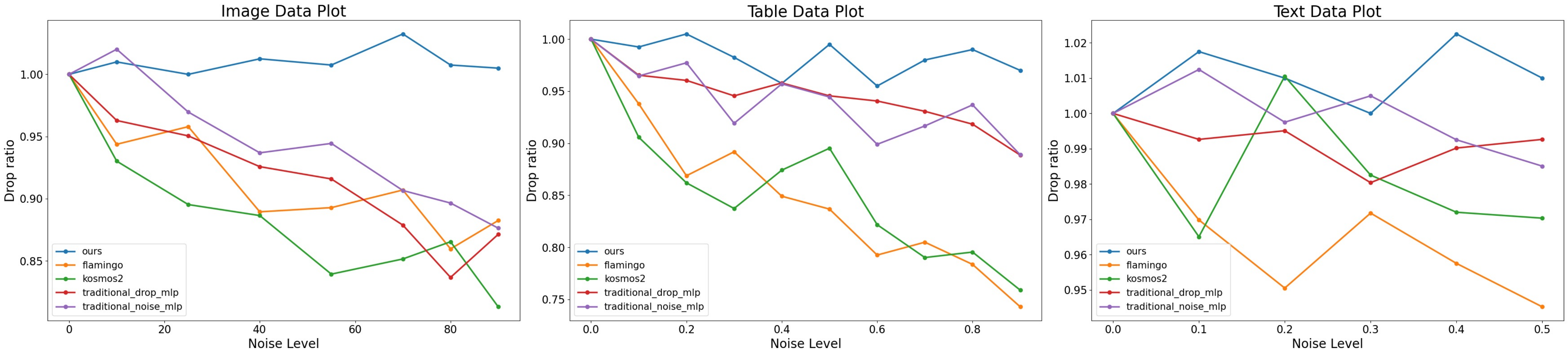}
  \caption{Drop ratio when noise is applied to modalities separately - Image (left) and Table (center) and Text (right).}
  \label{fig:acc_drop_image}
\end{figure*}

\end{document}